\documentclass[10pt,twocolumn,letterpaper]{article}

\usepackage[pagenumbers]{iccv} % To force page numbers, e.g. for an arXiv version

\usepackage{pifont}
\newcommand{\xmark}{\ding{55}}
\definecolor{iccvblue}{rgb}{0.21,0.49,0.74}
\usepackage[pagebackref,breaklinks,colorlinks,allcolors=iccvblue]{hyperref}
\usepackage[accsupp]{axessibility}

\def\paperID{14} % *** Enter the Paper ID here
\def\confName{ICCV DriveX Workshop}
\def\confYear{2025}

\title{Research Challenges and Progress in the End-to-End V2X \\ Cooperative Autonomous Driving Competition}

\author{
\textbf{Ruiyang Hao$^{1}$, Haibao Yu$^{2,1,}$\thanks{Corresponding author.}\;, Jiaru Zhong$^{1}$, Chuanye Wang$^{1}$, Jiahao Wang$^{1}$, } \\
\textbf{Yiming Kan$^{3}$, Wenxian Yang$^{1}$, Siqi Fan$^{1}$, Huilin Yin$^{3}$, Jianing Qiu$^{4}$, Yao Mu$^{2,5}$, } \\  
\textbf{Jiankai Sun$^{6}$, Li Chen$^{2,7}$, Walter Zimmer$^{8}$, Dandan Zhang$^{9}$, Shanghang Zhang$^{10}$, } \\
\textbf{Mac Schwager$^{6}$, Ping Luo$^{2,*}$, Zaiqing Nie$^{1,*}$} \\
$^{1}$ Tsinghua University,  $^{2}$ Hong Kong University,  $^{3}$ Tongji University, \\
$^{4}$ Chinese University of Hong Kong, $^{5}$Shanghai Jiao Tong University, $^{6}$Stanford University,  \\
$^{7}$OpenDriveLab, $^{8}$Technical University of Munich, $^{9}$Imperial College London, \\
$^{10}$Peking University}

\begin{document}
\maketitle
\begin{abstract}

With the rapid advancement of autonomous driving technology, vehicle-to-everything (V2X) communication has emerged as a key enabler for extending perception range and enhancing driving safety by providing visibility beyond the line of sight. However, integrating multi-source sensor data from both ego-vehicles and infrastructure under real-world constraints, such as limited communication bandwidth and dynamic environments, presents significant technical challenges.
To facilitate research in this area, we organized the End-to-End Autonomous Driving through V2X Cooperation Challenge, which features two tracks: cooperative temporal perception and cooperative end-to-end planning. Built on the UniV2X framework and the V2X-Seq-SPD dataset, the challenge attracted participation from over 30 teams worldwide and established a unified benchmark for evaluating cooperative driving systems.
This paper describes the design and outcomes of the challenge, highlights key research problems including bandwidth-aware fusion, robust multi-agent planning, and heterogeneous sensor integration, and analyzes emerging technical trends among top-performing solutions. By addressing practical constraints in communication and data fusion, the challenge contributes to the development of scalable and reliable V2X-cooperative autonomous driving systems.
\end{abstract}    
\section{Introduction}
\label{sec:intro}

Autonomous driving has witnessed rapid advancements in recent years, driven by the progress of perception~\cite{cai2021yolov4, chen2021deep}, planning~\cite{huang2023differentiable, huang2023gameformer}, and end-to-end~\cite{hu2023planning, jiang2023vad, chen2024end, wang2024driving, pan2024vlp} technologies. However, the prevailing paradigm of single-vehicle autonomy, which relies solely on onboard sensors and processing units, is inherently limited by its constrained field of view, susceptibility to occlusions, and lack of awareness of occluded or distant objects~\cite{bagheri20215g, yang2022autonomous}. These limitations pose significant challenges in complex urban environments, where safety-critical decision-making demands a more comprehensive understanding of the surrounding traffic context. In particular, scenarios involving intersections, occluded crosswalks, or multi-lane merges often expose the limitations of local perception and lead to suboptimal or unsafe maneuvers.

To address these constraints, vehicle-to-everything (V2X) cooperation has emerged as a promising paradigm~\cite{yusuf2024vehicle, 10715696}. By enabling ego-vehicles to exchange real-time sensory and state information with roadside infrastructure and nearby agents, V2X cooperation extends perception beyond the line of sight and supports more informed and robust perception and final planning performance~\cite{Song_2024_CVPR, 10588500}. The integration of cooperative perception and cooperative planning is thus becoming a pivotal frontier in the development of scalable and safe embodied intelligence systems for autonomous driving.

Despite the growing body of research on V2X-enabled systems, developing deployable and generalizable algorithms for cooperative driving remains challenging. Real-world constraints such as limited communication bandwidth~\cite{clancy2024wireless}, latency, and heterogeneous sensor configurations~\cite{xu2023cooperative} complicate the design of end-to-end solutions. Moreover, robust fusion of multi-view, multi-agent data~\cite{10720085, li2024di} for downstream planning under dynamic scenarios is still an open research problem. These challenges are further compounded by the asynchronous nature of inter-agent communication, variable sensor quality across nodes, and the lack of standardized protocols for representation and fusion.

To promote research in this direction, we organized the \textbf{\textit{first} End-to-End Autonomous Driving through V2X Cooperation Challenge} as part of the Multi-Agent Embodied Intelligent Systems (MEIS) Workshop @ CVPR 2025 (More details in this \href{https://coop-intelligence.github.io/V2X-Sec_MEIS/}{link}). The challenge aims to benchmark and advance the state-of-the-art in V2X-enhanced driving agents through two complementary tracks: (1) Cooperative Temporal Perception, focusing on multi-agent detection and tracking; and (2) Cooperative End-to-End Planning, targeting V2X-aware sensor-to-action learning. Built upon the open-source UniV2X framework~\cite{yu2025end} and V2X-Seq-SPD dataset~\cite{yu2023v2x}, this challenge provides a reproducible platform for evaluating cooperative perception and planning systems in real-world urban driving scenarios.

This paper presents a comprehensive summary of the competition design, research challenges, participant solutions, and key findings. Specifically, we (i) outline the motivation and structure of the challenge, (ii) identify critical research issues emerging from participant submissions, (iii) analyze the technical trends and progress demonstrated, and (iv) discuss future directions for cooperative multi-agent autonomous driving systems.

\begin{table*}[ht]
\centering
\caption{
Comparison of autonomous driving datasets by data source, held competitions, task description, V2X support, end-to-end (E2E) support.
\footnotesize{\textit{Abbreviations:} 
V2X = V2X model support,
E2E = End-to-End driving model support,
Det = Detection,
Trk = Tracking,
MPre = Motion Prediction,
Pla = Planning (Open-loop),
CL = Closed-loop evaluation
}
}
\begin{tabular}{lcccccc}
\hline
\hline
\textbf{Dataset} & \textbf{Reality} & \textbf{Competition} & \textbf{Task description} & \textbf{V2X} & \textbf{E2E} \\
\hline
nuScenes~\cite{caesar2020nuscenes} & Real   & CVPRW19, ICRAW20, ICRAW21           & Det,Trk,MPre,Pla    & \xmark  & \checkmark  \\
Waymo~\cite{sun2020scalability}   & Real     & WOD20-25         & Det,Trk,MPre,Pla             & \xmark  & \checkmark \\
Argoverse~\cite{chang2019argoverse}   & Real     & CVPRW22, CVPRW23, CVPRW25         & Det,Trk,MPre,Pla            &  \xmark  & \checkmark \\
CARLA~\cite{carla_leaderboard_2024}  & Sim     & CVPRW19, NIPSW20-22, CVPRW24         & Det,Trk,MPre,Pla,CL            &  \xmark  & \checkmark \\
NAVSIM~\cite{dauner2024navsim}  & Real     & CVPRW24, CVPRW25, ICCVW25         & Det,Trk,MPre,Pla,CL            &  \xmark  & \checkmark \\
DAIR-V2X~\cite{yu2022dair} & Real     & AIR-Apollo23         & Det            &  \checkmark  & \xmark \\
TUMTraf~\cite{zimmer2024tumtraf} & Real     & ICCVW25         & Det            &  \checkmark  & \xmark \\
V2v4Real~\cite{xu2023v2v4real} & Real     & --         & Det            &  \checkmark  & \xmark \\
V2X-Sim~\cite{li2022v2x} & Sim     & --         & Det,Trk            &  \checkmark  & \xmark \\
\hline
\textit{\textbf{V2X-Seq~\cite{yu2023v2x}}} & \textit{\textbf{Real}}     &  \textit{\textbf{CVPRW25 (Ours)}}         &  \textit{\textbf{Det,Trk,Pla}}            &   \textit{\textbf{\checkmark}}  & \textit{\textbf{\checkmark}} \\
\hline
\hline
\end{tabular}
\label{tab:dataset-comparison}
\end{table*}
\section{Background}
\label{sec:background}

\subsection{Related Benchmarks and Challenges}

Over the past decade, a variety of datasets and benchmarks have been proposed to evaluate the perception and planning capabilities of autonomous driving systems. Notable examples include nuScenes~\cite{caesar2020nuscenes}, Waymo Open Dataset~\cite{sun2020scalability}, Argoverse~\cite{chang2019argoverse}, nuplan-based dataset~\cite{caesar2021nuplan, peng2023openscene, dauner2024navsim, hao2025styledrive}, and the CARLA-based dataset~\cite{dosovitskiy2017carla, chitta2022transfuser, jia2024bench2drive}, which focus on object detection, motion prediction, and planning under the single-agent paradigm. While these benchmarks have significantly contributed to the development of perception, decision-making and end-to-end pipelines, they largely neglect the potential of inter-agent cooperation and V2X communication~\cite{yu2022dair, li2022v2x, zimmer2024tumtraf}, which are essential for overcoming occlusion and limited sensor range in congested urban environments. These limitations hinder the modeling of realistic traffic scenes involving multi-agent interactions and limited visibility, such as those found at intersections, curved roads, or occluded pedestrian zones.

Several recent efforts, such as DAIR-V2X~\cite{yu2022dair}, V2X-Sim~\cite{li2022v2x}, TUMTraf~\cite{zimmer2024tumtraf}, V2X-Real~\cite{xiang2024v2x}, V2v4Real~\cite{xu2023v2v4real}, RCooper\cite{hao2024rcooper}, Griffin\cite{wang2025griffin} and V2XSet~\cite{xu2022v2x}, have introduced datasets and tasks tailored for cooperative perception. These datasets incorporate multi-view inputs from vehicles and roadside infrastructure, enabling exploration of early and intermediate sensor fusion methods to enhance 3D detection and tracking performance. However, most of these benchmarks remain focused on perception tasks, with relatively limited emphasis on downstream planning~\cite{yu2023v2x}. In particular, few existing datasets provide a unified setting where both perception and planning tasks are evaluated with the same data and scenario structure.

The End-to-End V2X Cooperation Challenge addresses this gap by integrating cooperative perception and planning tasks into a two-track benchmark framework. It builds on the open-source UniV2X system~\cite{yu2025end} and the V2X-Seq-SPD dataset~\cite{yu2023v2x}, which jointly support detection, tracking, and motion planning based on multi-agent sensor inputs. By standardizing the task input/output formats and providing an end-to-end development pipeline, the challenge enables participants to explore perception-to-planning integration under realistic multi-view sensing conditions. The use of distinct sensing viewpoints and calibration setups naturally reflects challenges in real-world cooperative driving deployments. This joint benchmark structure promotes a more comprehensive understanding of algorithm performance in multi-agent urban environments.

\subsection{UniV2X Framework and Dataset}
The challenge is built upon the open-source UniV2X framework~\cite{yu2025end}, which serves as the first unified end-to-end pipeline for cooperative autonomous driving. UniV2X integrates multiple key modules—cooperative perception, intermediate representation learning, occupancy forecasting, and planning—into a cohesive architecture. It supports both vehicle-side and infrastructure-side sensing, facilitating multi-view feature alignment and fusion through a hybrid sparse-dense transmission protocol. This allows for efficient message passing while mitigating the communication burden common in dense feature maps, particularly in bird’s-eye-view (BEV) frameworks.

The underlying dataset, V2X-Seq-SPD~\cite{yu2023v2x}, provides synchronized and calibrated sensor recordings from ego vehicles and roadside units (RSUs), including front-view images, LiDAR point clouds (converted to BEV), and semantic commands. Ground-truth labels for 3D object detection, tracking, and future trajectories are included, allowing evaluation across both perception and planning tasks. The dataset reflects diverse urban driving scenarios with dynamic traffic flow, intersections, and occlusions—thus capturing key challenges faced by V2X systems.

\begin{figure*}[ht]
  \centering
  \begin{subfigure}{0.85\linewidth}
    \includegraphics[width=1.0\linewidth]{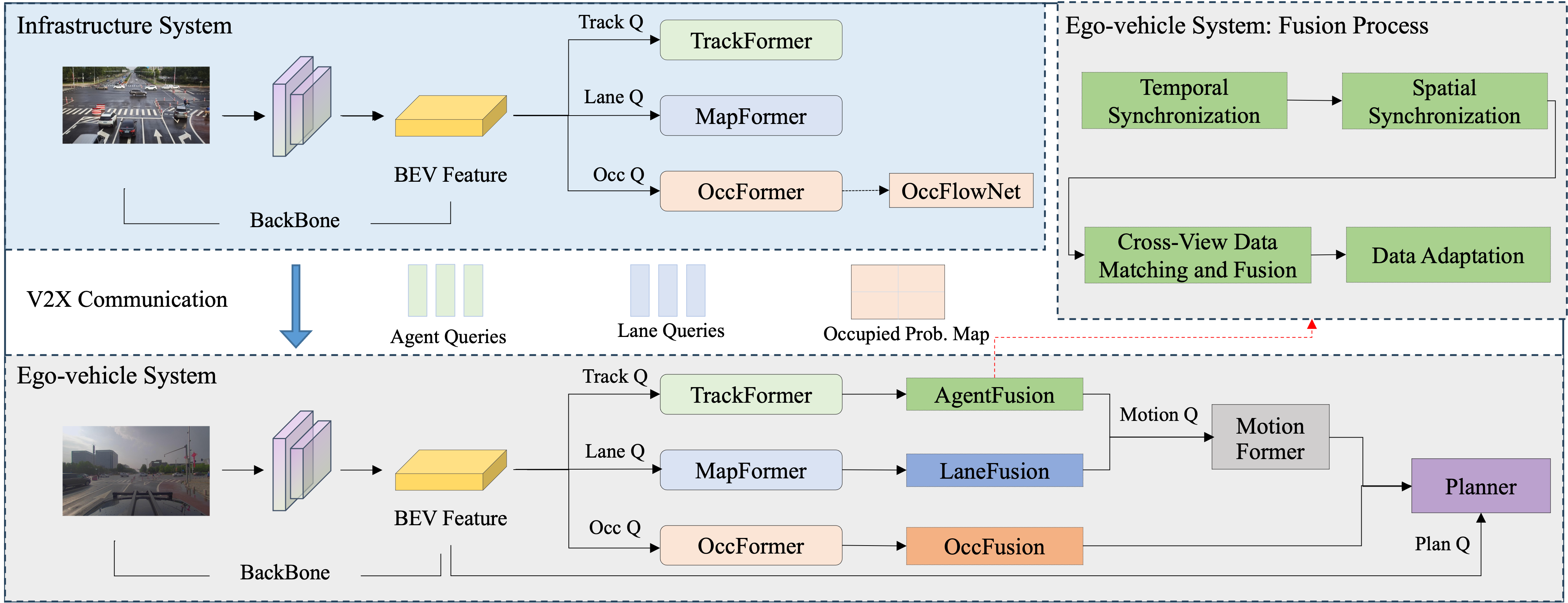}
    \caption{Challenge Baseline: UniV2X Architecture~\cite{yu2025end}.}
    \label{fig:short-a}
  \end{subfigure}
  \hfill
  \begin{subfigure}{0.85\linewidth}
    \includegraphics[width=1.0\linewidth]{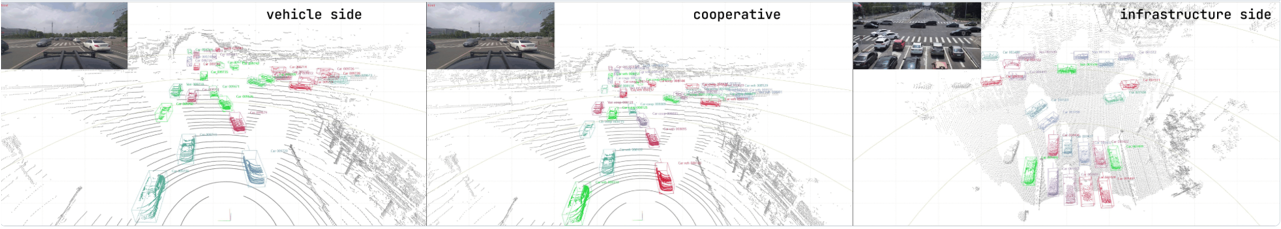}
    \caption{An example of V2X-Seq-SPD Dataset~\cite{yu2023v2x}}
    \label{fig:short-b}
  \end{subfigure}
  \caption{Challenge Baseline UniV2X~\cite{yu2025end} and V2X-Seq-SPD Dataset~\cite{yu2023v2x}}
  \label{fig:baseline_dataset}
\end{figure*}

UniV2X serves as the official baseline for both tracks of the competition. In Track 1, it provides a fully sparse 3D detection and tracking solution with anchor-guided query fusion. In Track 2, it offers a modular sensor-to-planning pipeline that leverages query-based adapters to dynamically route fused features into planning heads. These designs provide participants with a strong starting point and encourage innovation in overcoming current bottlenecks.
\section{Challenge Design}
\label{challenge_design}

\subsection{Task Setup and Evaluation Metrics}
The challenge comprises two complementary tracks designed to evaluate different aspects of V2X cooperative autonomous driving: Cooperative Temporal Perception and Cooperative End-to-End Planning.

\paragraph{1) Track 1: Cooperative Temporal Perception}
This track focuses on cooperative 3D detection and multi-object tracking in urban scenarios involving ego vehicles and roadside infrastructure. Each participant receives a stream of synchronized multi-agent sensor data, including front-view camera images from both ego vehicles and roadside units (RSUs), along with camera calibration parameters, vehicle ego states, and high-level command information. These inputs are drawn from realistic driving sequences, featuring intersections, dynamic obstacles, and partial observability across viewpoints.

The primary task is to detect vehicles of the merged “Car” category in 3D space and associate consistent tracking IDs across time, leveraging both temporal information and cross-agent collaboration. The design emphasizes the need for participants to model how complementary viewpoints—e.g., an RSU’s top-down view and the ego vehicle’s forward-facing camera—can be fused over time to disambiguate occluded or partially visible objects.

To evaluate performance, we employ two widely used metrics in cooperative perception benchmarks: mean Average Precision (mAP), which measures spatial detection accuracy, and Average Multi-Object Tracking Accuracy (AMOTA), which captures temporal consistency of object identities. The final evaluation score is computed as the unweighted average of the two (0.5 mAP + 0.5 AMOTA), allowing fair comparison between detection and tracking capabilities.

This task encourages the design of fusion algorithms capable of aligning features from spatially distinct viewpoints and maintaining identity consistency across frames, even under object occlusion, motion blur, or disjoint agent fields of view. It also offers a platform to evaluate temporal modeling techniques such as query-based memory propagation, agent-aware attention, and cross-frame association strategies. Ultimately, this track aims to advance the robustness and scalability of cooperative perception systems deployed in real-world driving environments.

\paragraph{2) Track 2: Cooperative End-to-End Planning}
This track aims to evaluate complete sensor-to-planning pipelines that generate future motion trajectories based on fused perception from multiple agents. Participants are tasked with predicting a sequence of future waypoints over a 5-second horizon, using the same input modalities as in Track 1, including ego and infrastructure camera images, calibration data, command signals, and current ego vehicle states.

Unlike modular approaches that decouple perception and planning, this track encourages joint reasoning across the full autonomous driving stack, from raw sensor input to trajectory-level output. The data spans a variety of challenging urban situations—such as intersection negotiation, overtaking, and lane turning—requiring the agent to anticipate dynamic scene evolution and react safely under partial observability.

Performance is assessed using three complementary metrics:
\begin{itemize}
    \item L2 Error, which measures the Euclidean distance between predicted and ground-truth waypoints, reflecting trajectory accuracy;
    \item Collision Rate, which quantifies how often the predicted trajectory intersects with other traffic participants;
    \item Off-road Rate, which measures deviation from the drivable area and thus reflects constraint violation or poor lane adherence.
\end{itemize}

To obtain a comprehensive evaluation, each metric is averaged at three future timestamps (2.5s, 3.5s, 4.5s), balancing short-term responsiveness and long-term planning quality. A min-max normalization is applied based on predefined reference ranges, and the final score is computed as a weighted sum:
0.5 × normalized L2 Error + 0.25 × normalized Collision Rate + 0.25 × normalized Off-road Rate.

This track emphasizes planning robustness in complex multi-agent scenes, and highlights the importance of integrating spatial-temporal reasoning, intent understanding, and safety guarantees into the learning process. It offers a testbed for evaluating architectures such as transformer-based fusion planners, modular policy networks, and multi-head decoding strategies under realistic traffic conditions.

\subsection{Participation}

Over 30 teams registered, with 5 finalists achieving ranked results. Participants came from academic institutions and industry research labs across China, Japan, the Middle East, the United States, and Europe. Most teams adopted the open-source UniV2X baseline as a foundation, developing innovative fusion architectures and planning strategies on top of it. To recognize outstanding solutions, the challenge organizers awarded monetary prizes to the top-ranked teams in each track. The diversity in approaches—from sparse query-based perception pipelines to modular planning frameworks—reflects the richness and complexity of the V2X cooperation landscape.
\section{Research Challenges}
\label{research_challenges}

The V2X Cooperation Challenge was intentionally designed to reflect real-world difficulties in cooperative autonomous driving. Through analysis of participant submissions and related work, several core research challenges emerged, spanning multi-agent fusion, communication efficiency, planning robustness, and realistic deployment modeling. These challenges reveal both the current limitations of existing solutions and promising directions for future research.

\paragraph{Multi-Agent Sensor Fusion under Bandwidth Constraints.}
A fundamental challenge lies in effectively aggregating heterogeneous sensor inputs from ego vehicles and infrastructure, particularly under tight communication budgets. Naïvely transmitting dense feature maps from multiple viewpoints (e.g., bird’s-eye view or BEV) quickly exhausts bandwidth and leads to latency bottlenecks~\cite{chang2023bev, yu2023flow}. More recent methods employ sparse query-based methods and transformer for cooperative representations embedding and fusion~\cite{fan2024quest, zhong2024leveraging, wang2025coopdetr, zhong2025cooptrack}. This necessitates the development of sparse, information-aware representations that can preserve critical scene understanding while minimizing message size. 

Top-performing teams in Track 1 adopted query-based attention fusion mechanisms, such as anchor-guided sparse queries and cooperative instance denoising, to mitigate these issues. However, challenges remain in dynamically selecting which information to transmit, how to encode uncertainty from partial observations, and how to align features from spatially and temporally misaligned views. Efficient and adaptive feature compression strategies, potentially guided by learned importance scores, are still underexplored.

\paragraph{Robust Planning in Dynamic and Complex Environments}
Track 2 highlighted the difficulty of producing reliable motion plans in highly dynamic, multi-agent urban scenes. When relying on fused perception from multiple sources, temporal inconsistency, latency-induced misalignment, and partial observability can significantly degrade planning performance~\cite{xiang2025v2x, zhao2024remote}. Ego agents must reason not only about static obstacles and drivable regions, but also about the future intentions and potential interactions of nearby vehicles.

Moreover, the planning module must cope with command diversity (e.g., turns, stops, merges) and structural uncertainty in intersections or occluded traffic elements. These issues call for more robust multi-modal trajectory prediction, tighter integration of intent inference, and online failure recovery mechanisms in planning architectures.

\paragraph{Communication-Aware System Design and Modeling}
Realistic V2X deployment is subject to a range of networking imperfections, including packet loss~\cite{mouawad2021collective}, varying latency, and intermittent connectivity~\cite{ren2024interruption}. However, most existing cooperative driving methods assume idealized or fixed-delay channels~\cite{yu2025end, you2024v2x}. The challenge dataset incorporates limited communication constraints (e.g., message size limits), but further progress depends on building systems that are explicitly aware of and adaptive to the communication channel.

Few teams explored bandwidth-adaptive fusion strategies or uncertainty-aware planning under degraded connectivity. Future systems can reason about when, what, and how to communicate, potentially leveraging learned policies or information-theoretic objectives. Modeling the trade-off between perception gain and communication cost remains an open research question, especially when agents must operate asynchronously or with partial participation.

\paragraph{Generalization and Transfer under Domain Shift}
Although the dataset provides consistent sensor configurations, real-world deployments often involve heterogeneous sensor suites, diverse camera placements, and varying calibration quality~\cite{zhao2024coopre, zha2025heterogeneous}. Designing fusion and planning models that generalize across these variations remains challenging. Furthermore, reliance on known object models or tightly coupled training scenarios can hinder transferability to new domains.

Some participants addressed this by employing modular architectures with adaptable feature backbones, but the issue of domain robustness under limited supervision persists. Robustness to weather, lighting, and sensor degradation was not evaluated in this challenge but constitutes a necessary extension for real-world readiness.

\section{Progress and Analysis}
\label{progress_analysis}

The competition attracted a diverse set of participants from academia and industry, contributing a broad spectrum of approaches across cooperative perception, feature fusion, and planning architectures. While implementations varied in complexity and formulation, a number of converging trends emerged. In particular, the most effective solutions reflect a growing shift toward modular, interpretable, and task-centric designs that emphasize structured information flow between agents and system components.

This section introduces the top-performing solutions from each track of the challenge. These methods represent state-of-the-art approaches in cooperative 3D perception and end-to-end planning with V2X input, and demonstrate the effectiveness of structured representations and adaptive fusion strategies.

\subsection{Track 1 Top Method: SparseCoop}
\begin{figure*}[ht]
  \centering
  \includegraphics[width=0.9\linewidth]{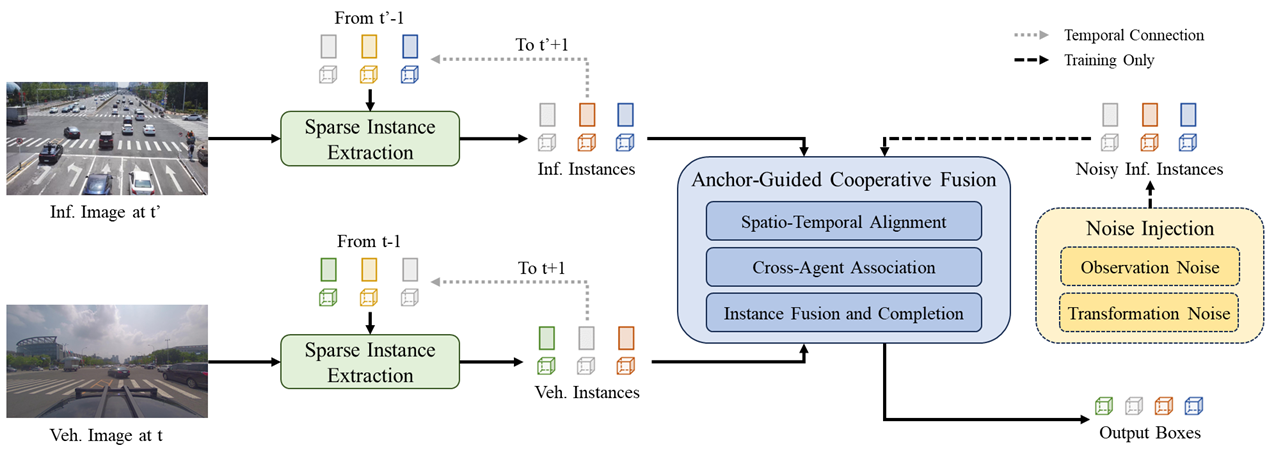}
  \caption{Architecture of \textbf{SparseCoop}, the top-ranked solution in \textbf{Track 1: Cooperative Temporal Perception}. The method adopts a fully sparse cooperative detection and tracking pipeline, where each object is represented by an \textit{anchor-aided instance query} containing structured geometric attributes (position, size, velocity, orientation) and semantic features. Cross-agent fusion is performed directly at the object level without relying on intermediate BEV features. A \textit{cooperative instance denoising task} is applied during training to inject noise into ground-truth anchors and improve convergence robustness through reconstruction supervision.}
  \label{fig:sparsecoop}
\end{figure*}

Wang et al. from Tsinghua University proposed SparseCoop, a fully sparse, instance-centric cooperative perception framework (Fig.~\ref{fig:sparsecoop}) designed to simultaneously address the communication and computational bottlenecks of traditional dense BEV-based approaches and the challenges of newer sparse, query-based methods, including their insufficiently expressive query representations for handling real-world scenarios and their inherent training instability.

At its core, SparseCoop introduces the concept of the anchor-aided instance query, where each object is represented by a rich feature vector coupled with an explicit anchor box. The anchor includes structured geometric and motion attributes—namely the object’s 3D position, dimensions, velocity, and yaw. This representation enables precise, physically grounded fusion across agents with different viewpoints and asynchronous observations.

To address the training instability common in sparse query systems, SparseCoop incorporates a cooperative instance denoising task. During training, noise is deliberately added to ground-truth objects in the form of "Observation Noise" and "Transformation Noise". The model is then supervised to recover clean object states, which generates a robust and abundant stream of positive training signals. This design improves convergence speed and accuracy.

SparseCoop achieves state-of-the-art detection and tracking performance, demonstrating strong robustness to viewpoint diversity, temporal misalignment, and perception noise under the V2X-Seq-SPD benchmark.

\subsection{Track 2 Top Method: MAP}
\begin{figure*}[ht]
  \centering
  \includegraphics[width=0.8\linewidth]{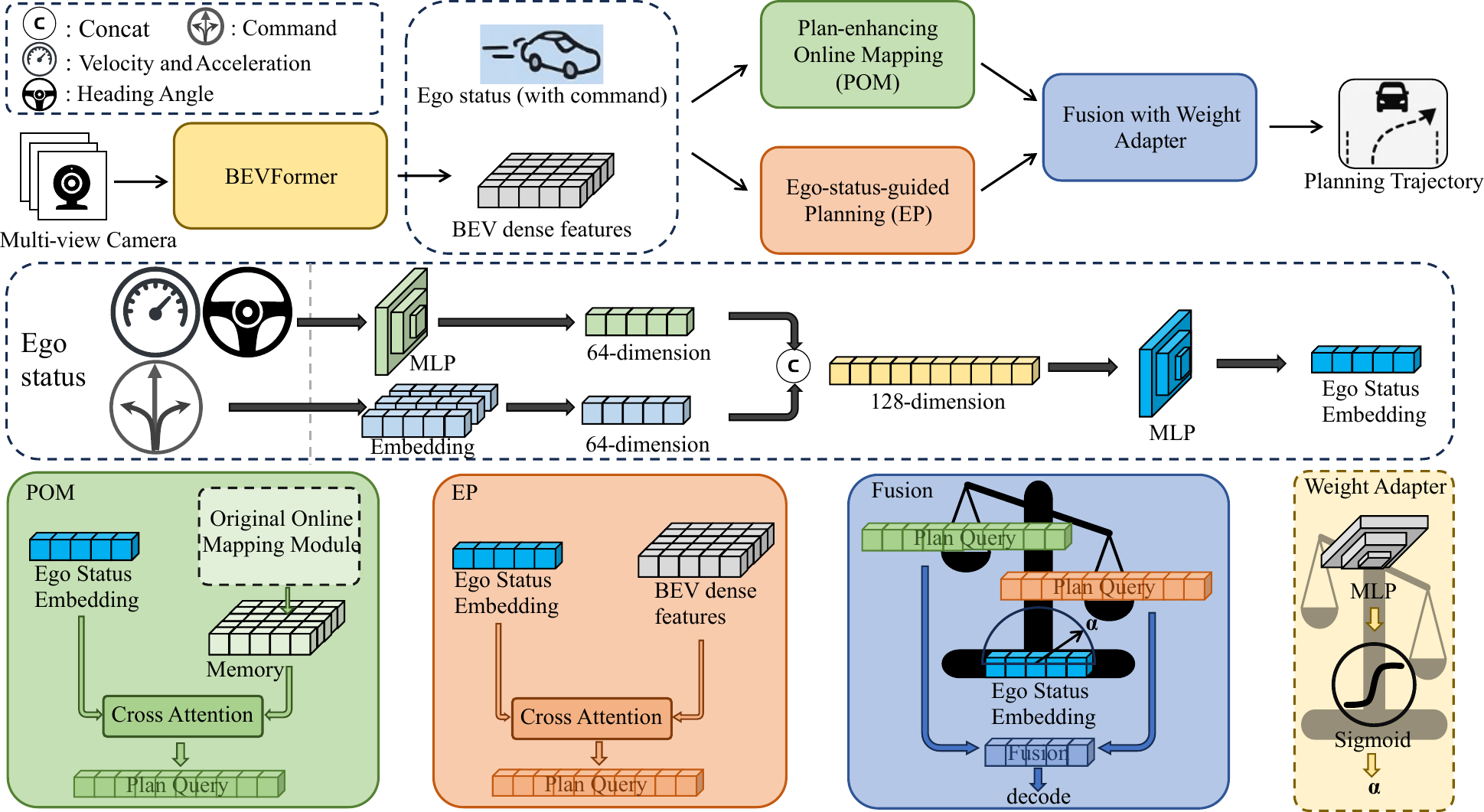}
  \caption{Architecture of \textbf{MAP}, the top-ranked solution in \textbf{Track 2: Cooperative End-to-End Planning}. This planning-centric framework explicitly incorporates semantic map information into trajectory generation. It consists of two query-generation branches: the \textit{Plan-enhancing Online Mapping (POM)} module extracts semantic priors from segmentation outputs, while the \textit{Ego-status-guided Planning (EP)} module models motion dynamics. A learned adapter fuses the two planning queries with weights conditioned on ego state, enabling context-aware trajectory generation under varying traffic complexity.}
  \label{fig:map}
\end{figure*}

The MAP framework (Fig.~\ref{fig:map}), proposed by Kan et al. from Tongji University, emerged from a critical reevaluation of the role of perception in end-to-end autonomous driving. While many recent approaches favor minimal input paradigms that rely solely on ego history, MAP challenges this trend by demonstrating that explicitly and effectively utilizing semantic map information can substantially enhance planning robustness.

At its core, MAP transforms semantic segmentation from a passive supervision target into a direct planning input. It introduces a two-branch query generation pipeline: The Ego-status-guided Planning (EP) module leverages the current ego state for trajectory planning, while the other extracts map-guided priors through a Plan-enhancing Online Mapping (POM) module. The resulting semantic-aware and ego-status-driven queries are then fused via a learned Weight Adapter, which adaptively predicts a fusion scalar $\alpha$ based on the current driving context.

This adaptive weighting mechanism allows the planner to rely more on ego information in simple scenes, and to prioritize semantic priors in complex or ambiguous scenarios, leading to context-sensitive and reliable decision-making. Importantly, MAP achieves strong performance without stacked modules such as tracking or occupancy prediction.

On the DAIR-V2X-Seq-SPD benchmark, MAP improves the overall normalized score by 44.5\% over the UniV2X baseline and ranks first on the planning leaderboard, showing competitive results across all sub-metrics, including L2 error and off-road rate.

\section{Future Directions}
\label{future_directions}

The challenge results and observed limitations across both tracks reveal a clear gap between current benchmark performance and the requirements for real-world deployment of cooperative autonomous driving under V2X settings. To close this gap, future research should address the following interconnected directions, progressing from foundational communication challenges to system-level adoption.

\subsection{Realistic V2X Communication Modeling}
Most current solutions assume ideal or simplified communication channels with perfect message delivery~\cite{xu2022v2x, you2024v2x}. In practice, V2X networks suffer from variable latency, intermittent connectivity, and packet drops caused by interference or congestion~\cite{coll2022end, souri2024systematic}. Future benchmarks and algorithms should embed communication-aware learning by:
\begin{itemize}
    \item Simulating packet loss models grounded in empirical wireless measurements~\cite{gularte2024integrating},
    \item Incorporating delay-aware fusion mechanisms that reason with stale or missing data~\cite{ahmed2025delawarecol, xu2024delay},
    \item Designing redundancy-aware protocols or modules to prioritize safety-critical and planning-oriented information under bandwidth constraints~\cite{hui2023rcfl}.
\end{itemize}

These modules would enable robust agents that adapt to both perceptual uncertainty and communication reliability.

\subsection{Bandwidth-Adaptive and Task-Aware Fusion}
Fusion strategies must adapt to both bandwidth availability and task requirements to ensure reliable performance at scale~\cite{yu2023flow}. However, most existing approaches remain static and fail to capture the dynamic trade-offs between efficiency and accuracy. Beyond sparse fusion, future feature fusion modules may benefit from:
\begin{itemize}
    \item Information-theoretic feature selection to maximize task utility per transmitted bit~\cite{ma2024class, chen2024effective},
    \item Hierarchical encoding schemes for coarse-to-fine updates based on link conditions~\cite{li2024coarse, luo2024full},
    \item Task-driven prioritization, sending safety-critical and planning-critical cues (e.g., nearby dynamic agents) more aggressively than static context.
\end{itemize}

These adaptive fusion mechanisms would support graceful degradation and efficient resource utilization in large-scale V2X deployments.

\subsection{Generalization Across Heterogeneous Agents and Scenarios}
Real-world deployments will involve heterogeneous vehicles and infrastructure with varying sensors, fields of view, and computational capabilities~\cite{zha2025heterogeneous, wang2025v2x, wei2025hecofuse}. Robust fusion and planning under such diversity can be supported by:
\begin{itemize}
    \item Calibration-agnostic fusion mechanisms or frameworks tolerant to partial or inaccurate alignment~\cite{fu2025self},
    \item Meta-learning or domain adaptation methods to generalize across sensor setups, cities, and conditions~\cite{li2024v2x, kong2023dusa},
    \item Scalable fusion topologies that support dynamic participation as agents enter or leave the scene~\cite{tan2024dynamic}.
\end{itemize}

Addressing these challenges will significantly improve the deployability of cooperative driving systems across different geographies and manufacturers.

\subsection{Air-Ground Collaboration}
The growth of the low-altitude economy~\cite{wang2025toward} will introduce drones as additional sensing and communication agents for autonomous driving~\cite{gao2025airv2x, wang2025griffin, hou2025agc, feng2024u2udata}. By acting as “free viewpoints,” drones can enrich situational awareness in dense traffic and occluded intersections. At the same time, their deployment raises issues of sensor vibration, communication overhead, and limited endurance. Air-ground cooperative systems can be advanced by:
\begin{itemize}
    \item Addressing sensor vibration artifacts in drone-mounted perception, which can degrade image quality and downstream detection and planning performance~\cite{11006752},
    \item Developing bandwidth-aware fusion algorithms for transmitting aerial data efficiently,
    \item Coordinating multiple drones under battery and flight-time constraints for persistent coverage~\cite{chen2024fast, chughtai2024drone}.
\end{itemize}

Leveraging aerial viewpoints in coordination with ground vehicles could unlock richer situational awareness and enable safer, large-scale deployment of cooperative driving systems.

\subsection{Interpretability, Safety, and Standardization}
For cooperative systems to be adopted in safety-critical applications such as autonomous driving, interpretability, verifiability and standardization become essential~\cite{lei2025risk, yu2025end, ali2025vehicles}. Progress can be made through:
\begin{itemize}
    \item Transparent fusion architectures that expose the contribution of each agent and observation~\cite{10720085, lei2025risk},
    \item Uncertainty quantification in cooperative perception and planning outputs~\cite{li2025efficient,10618989},
    \item Conformance to communication and safety standards such as SAE~J2735~\cite{sumner2013sae} or ETSI~ITS-G5~\cite{vogt2024comprehensive}.
\end{itemize}

These directions are critical to ensuring that cooperative driving systems are not only performant but also trustworthy and certifiable for real-world use.

% \subsection{Language as Communication Medium}
% Recent advances in Large Vision-Language Models~\cite{jiang2025survey, xing2024autotrust, wang2025generative} suggest natural language as an alternative medium for V2X communication~\cite{gao2025langcoop, gao2025automated, cui2025towards, luo2025v2x, wu2025v2x, chiu2025v2v}. Compared to transmitting raw sensor data or neural features, natural language enables:
% \begin{itemize}
%     \item Greater transparency and explainability of cooperative decisions~\cite{gao2025automated},
%     \item Lower bandwidth requirements relative to high-dimensional sensor streams~\cite{gao2025langcoop},
%     \item Model-agnostic interoperability across heterogeneous agents~\cite{gao2025stamp, xia2025one},
%     \item Intent-and-decision-level negotiation between vehicles and even pedestrians~\cite{gao2025automated, cui2025towards}.
% \end{itemize}

% Exploring language as a communication medium may open up new paradigms for intent sharing and human-AI interaction in cooperative driving.

\subsection{Language as Communication Medium}
Recent advances in Large Vision-Language Models~\cite{jiang2025survey, xing2024autotrust, wang2025generative} have opened the possibility of using natural language as a medium for V2X communication~\cite{gao2025langcoop, gao2025automated, cui2025towards, luo2025v2x, wu2025v2x, chiu2025v2v, gao2025stamp, xia2025one}. Language-based communication promises transparency, efficiency, and interoperability, but its deployment in safety-critical driving scenarios remains largely unexplored. Future research should address:
\begin{itemize}
    \item Developing structured and unambiguous protocols for language-based V2X exchanges to avoid ambiguity~\cite{gao2025automated, wu2025v2x},
    \item Combining natural language with traditional feature-level or state-level communication in hybrid pipelines,
    \item Studying robustness to multilingual, noisy, or adversarial language inputs in cooperative driving,
    \item Exploring decision-level negotiation mechanisms that go beyond perception sharing~\cite{cui2025vehicle}.
\end{itemize}

Addressing these challenges would transform natural language from a promising idea into a practical medium for cooperative autonomous driving.

\subsection{Community and Ecosystem Development}
Progress in V2X cooperative driving will be accelerated by cohesive community efforts and shared infrastructure. Key steps include:
\begin{itemize}
    \item Continuing development of open-source toolkits, such as UniV2X, for full-stack experimentation,
    \item Expanding datasets to cover adverse conditions (e.g., night, rain, sensor failures),
    \item Establishing long-term multi-institutional benchmarks to ensure reproducibility and collaboration,
    \item Organizing V2X-specific competitions to stimulate innovation beyond single-agent autonomy.
    \item Considering broader ethical and policy implications of multi-agent cooperation, including inter-manufacturer trust and data governance among automotive original equipment manufacturers (OEMs), as well as data security and regulatory compliance.
\end{itemize}

Building such community resources and ecosystems will help translate academic advances into robust, deployable cooperative driving systems worldwide.
\section{Conclusion}
\label{conclusion}

This paper has presented a comprehensive overview of the End-to-End Autonomous Driving through V2X Cooperation Challenge, organized as part of the MEIS Workshop @ CVPR 2025. The challenge was designed to advance the state of cooperative autonomous driving by rigorously evaluating perception and planning systems under realistic multi-agent and communication-constrained conditions. It comprised two tracks: cooperative temporal perception and end-to-end planning, built upon the open-source UniV2X framework~\cite{yu2025end} and the V2X-Seq-SPD dataset~\cite{yu2023v2x}.

Participation from numerous teams highlighted both substantial progress and persistent limitations in V2X-enabled driving systems. The strongest submissions adopted sparse, query-based fusion, modular architectures, and temporal reasoning, achieving competitive results in both perception and planning tasks. At the same time, we summarize the critical open challenges, including communication-aware fusion, robust planning under partial observability, and generalization across heterogeneous agents.

These findings and insights from related research, emphasize that the development of cooperative driving systems must go beyond accuracy and efficiency, extending to adaptability, interpretability, and robustness in real-world deployments. The challenge has further underscored the importance of open benchmarks, reproducible baselines, and cross-community collaboration as essential drivers for translating academic innovation into deployable systems.

Looking forward, future editions of the challenge will broaden their scope to incorporate richer sensor modalities, more realistic and dynamic communication models, and increasingly diverse driving environments. By continuing to integrate technical advances with ethical, regulatory, and societal considerations, this initiative seeks to foster the design of safe, scalable, and intelligent multi-agent driving systems for the urban mobility ecosystems of tomorrow.

\section*{Acknowledgement}
\label{ack}
The authors would like to express their sincere gratitude to all participating teams of our challenge for their valuable contributions. In particular, we acknowledge the outstanding efforts of the team led by Ziyi Song and Dr. Sheng Zhou from Tsinghua University, as well as the team led by Dr. Ehsan Javanmardi from the University of Tokyo. We would also like to thank Xiangbo Gao and Dr. Zhengzhong Tu from Texas A\&M University for their valuable suggestions on future work. We also gratefully acknowledge the sponsorship provided by the Multimedia Laboratory at the University of Hong Kong and Shanghai Songying Technology Co., Ltd. Furthermore, we sincerely acknowledge the support from the Wuxi Research Institute of Applied Technologies at Tsinghua University under Grant No. 20242001120.
{
    \small
    \bibliographystyle{ieeenat_fullname}
    \bibliography{main}

\begin{thebibliography}{94}
\providecommand{\natexlab}[1]{#1}
\providecommand{\url}[1]{\texttt{#1}}
\expandafter\ifx\csname urlstyle\endcsname\relax
  \providecommand{\doi}[1]{doi: #1}\else
  \providecommand{\doi}{doi: \begingroup \urlstyle{rm}\Url}\fi

\bibitem[Ahmed et~al.(2025)Ahmed, Mercelis, and Anwar]{ahmed2025delawarecol}
Ahmed~N. Ahmed, Siegfried Mercelis, and Ali Anwar.
\newblock Delawarecol: Delay aware collaborative perception.
\newblock \emph{IEEE Open Journal of Vehicular Technology}, 6:\penalty0 1164--1177, 2025.

\bibitem[Ali et~al.(2025)Ali, Nauman, Jamshed, Kim, and Kim]{ali2025vehicles}
Moin Ali, Ali Nauman, Muhammad~Ali Jamshed, Su~Min Kim, and Junsu Kim.
\newblock Vehicles-to-everything standardization, services and enhancements for intelligent transportation systems.
\newblock \emph{IEEE Communications Standards Magazine}, 2025.

\bibitem[Bagheri et~al.(2021)Bagheri, Noor-A-Rahim, Liu, Lee, Pesch, Moessner, and Xiao]{bagheri20215g}
Hamidreza Bagheri, Md Noor-A-Rahim, Zilong Liu, Haeyoung Lee, Dirk Pesch, Klaus Moessner, and Pei Xiao.
\newblock 5g nr-v2x: Toward connected and cooperative autonomous driving.
\newblock \emph{IEEE Communications Standards Magazine}, 5\penalty0 (1):\penalty0 48--54, 2021.

\bibitem[Bertocco et~al.(2025)Bertocco, Brighente, Ciattaglia, Gambi, Peruzzi, Pozzebon, and Spinsante]{11006752}
Matteo Bertocco, Alessandro Brighente, Gianluca Ciattaglia, Ennio Gambi, Giacomo Peruzzi, Alessandro Pozzebon, and Susanna Spinsante.
\newblock Malicious drone identification by vibration signature measurement: A radar-based approach.
\newblock \emph{IEEE Transactions on Instrumentation and Measurement}, 74:\penalty0 8004415, 2025.

\bibitem[Caesar et~al.(2020)Caesar, Bankiti, Lang, Vora, Liong, Xu, Krishnan, Pan, Baldan, and Beijbom]{caesar2020nuscenes}
Holger Caesar, Varun Bankiti, Alex~H Lang, Sourabh Vora, Venice~Erin Liong, Qiang Xu, Anush Krishnan, Yu Pan, Giancarlo Baldan, and Oscar Beijbom.
\newblock nuscenes: A multimodal dataset for autonomous driving.
\newblock In \emph{Proceedings of the IEEE/CVF conference on computer vision and pattern recognition}, pages 11621--11631, 2020.

\bibitem[Caesar et~al.(2021)Caesar, Kabzan, Tan, Fong, Wolff, Lang, Fletcher, Beijbom, and Omari]{caesar2021nuplan}
Holger Caesar, Juraj Kabzan, Kok~Seang Tan, Whye~Kit Fong, Eric Wolff, Alex Lang, Luke Fletcher, Oscar Beijbom, and Sammy Omari.
\newblock nuplan: A closed-loop ml-based planning benchmark for autonomous vehicles.
\newblock \emph{arXiv preprint arXiv:2106.11810}, 2021.

\bibitem[Cai et~al.(2024)Cai, Qu, Liu, Chen, and Xie]{10618989}
Kunyang Cai, Ting Qu, Fen Liu, Hong Chen, and Lihua Xie.
\newblock Cooperative perception with localization uncertainty: A cubature split covariance intersection framework.
\newblock \emph{IEEE Transactions on Intelligent Transportation Systems}, 25\penalty0 (11):\penalty0 18006--18024, 2024.

\bibitem[Cai et~al.(2021)Cai, Luan, Gao, Wang, Chen, Li, Sotelo, and Li]{cai2021yolov4}
Yingfeng Cai, Tianyu Luan, Hongbo Gao, Hai Wang, Long Chen, Yicheng Li, Miguel~Angel Sotelo, and Zhixiong Li.
\newblock Yolov4-5d: An effective and efficient object detector for autonomous driving.
\newblock \emph{IEEE Transactions on Instrumentation and Measurement}, 70:\penalty0 1--13, 2021.

\bibitem[Chang et~al.(2023)Chang, Zhang, Zhang, Zhong, Peng, Li, and Li]{chang2023bev}
Cheng Chang, Jiawei Zhang, Kunpeng Zhang, Wenqin Zhong, Xinyu Peng, Shen Li, and Li Li.
\newblock Bev-v2x: Cooperative birds-eye-view fusion and grid occupancy prediction via v2x-based data sharing.
\newblock \emph{IEEE Transactions on Intelligent Vehicles}, 8\penalty0 (11):\penalty0 4498--4514, 2023.

\bibitem[Chang et~al.(2019)Chang, Lambert, Sangkloy, Singh, Bak, Hartnett, Wang, Carr, Lucey, Ramanan, et~al.]{chang2019argoverse}
Ming-Fang Chang, John Lambert, Patsorn Sangkloy, Jagjeet Singh, Slawomir Bak, Andrew Hartnett, De Wang, Peter Carr, Simon Lucey, Deva Ramanan, et~al.
\newblock Argoverse: 3d tracking and forecasting with rich maps.
\newblock In \emph{Proceedings of the IEEE/CVF conference on computer vision and pattern recognition}, pages 8748--8757, 2019.

\bibitem[Chen et~al.(2021)Chen, Lin, Lu, Cao, Wu, Guo, Liu, and Wang]{chen2021deep}
Long Chen, Shaobo Lin, Xiankai Lu, Dongpu Cao, Hangbin Wu, Chi Guo, Chun Liu, and Fei-Yue Wang.
\newblock Deep neural network based vehicle and pedestrian detection for autonomous driving: A survey.
\newblock \emph{IEEE Transactions on Intelligent Transportation Systems}, 22\penalty0 (6):\penalty0 3234--3246, 2021.

\bibitem[Chen et~al.(2024{\natexlab{a}})Chen, Wu, Chitta, Jaeger, Geiger, and Li]{chen2024end}
Li Chen, Penghao Wu, Kashyap Chitta, Bernhard Jaeger, Andreas Geiger, and Hongyang Li.
\newblock End-to-end autonomous driving: Challenges and frontiers.
\newblock \emph{IEEE Transactions on Pattern Analysis and Machine Intelligence}, 46\penalty0 (12):\penalty0 10164--10183, 2024{\natexlab{a}}.

\bibitem[Chen et~al.(2024{\natexlab{b}})Chen, Zhu, Liu, and Guo]{chen2024fast}
Wu Chen, Jiayi Zhu, Jiajia Liu, and Hongzhi Guo.
\newblock A fast coordination approach for large-scale drone swarm.
\newblock \emph{Journal of Network and Computer Applications}, 221:\penalty0 103769, 2024{\natexlab{b}}.

\bibitem[Chen and Wang(2024)]{chen2024effective}
Yihao Chen and Zefang Wang.
\newblock An effective information theoretic framework for channel pruning.
\newblock \emph{arXiv preprint arXiv:2408.16772}, 2024.

\bibitem[Chitta et~al.(2022)Chitta, Prakash, Jaeger, Yu, Renz, and Geiger]{chitta2022transfuser}
Kashyap Chitta, Aditya Prakash, Bernhard Jaeger, Zehao Yu, Katrin Renz, and Andreas Geiger.
\newblock Transfuser: Imitation with transformer-based sensor fusion for autonomous driving.
\newblock \emph{IEEE transactions on pattern analysis and machine intelligence}, 45\penalty0 (11):\penalty0 12878--12895, 2022.

\bibitem[Chiu et~al.(2025)Chiu, Hachiuma, Wang, Smith, Wang, and Chen]{chiu2025v2v}
Hsu-kuang Chiu, Ryo Hachiuma, Chien-Yi Wang, Stephen~F Smith, Yu-Chiang~Frank Wang, and Min-Hung Chen.
\newblock V2v-llm: Vehicle-to-vehicle cooperative autonomous driving with multi-modal large language models.
\newblock \emph{arXiv preprint arXiv:2502.09980}, 2025.

\bibitem[Chughtai et~al.(2024)Chughtai, Qadri, Kaleem, and Yuen]{chughtai2024drone}
Omer Chughtai, Nadia~Nawaz Qadri, Zeeshan Kaleem, and Chau Yuen.
\newblock Drone-assisted cooperative routing scheme for seamless connectivity in v2x communication.
\newblock \emph{IEEE Access}, 12:\penalty0 17369--17381, 2024.

\bibitem[Clancy et~al.(2024)Clancy, Mullins, Deegan, Horgan, Ward, Eising, Denny, Jones, and Glavin]{clancy2024wireless}
Joseph Clancy, Darragh Mullins, Brian Deegan, Jonathan Horgan, Enda Ward, Ciar{\'a}n Eising, Patrick Denny, Edward Jones, and Martin Glavin.
\newblock Wireless access for v2x communications: Research, challenges and opportunities.
\newblock \emph{IEEE Communications Surveys \& Tutorials}, 26\penalty0 (3):\penalty0 2082--2119, 2024.

\bibitem[Coll-Perales et~al.(2022)Coll-Perales, Lucas-Esta{\~n}, Shimizu, Gozalvez, Higuchi, Avedisov, Altintas, and Sepulcre]{coll2022end}
Baldomero Coll-Perales, M~Carmen Lucas-Esta{\~n}, Takayuki Shimizu, Javier Gozalvez, Takamasa Higuchi, Sergei Avedisov, Onur Altintas, and Miguel Sepulcre.
\newblock End-to-end v2x latency modeling and analysis in 5g networks.
\newblock \emph{IEEE Transactions on Vehicular Technology}, 72\penalty0 (4):\penalty0 5094--5109, 2022.

\bibitem[{Contributors}(2024)]{carla_leaderboard_2024}
{Contributors}.
\newblock Carla autonomous driving leaderboard, 2024.

\bibitem[Cui et~al.(2025{\natexlab{a}})Cui, Tang, Holtz, Nguyen, Allievi, Qiu, and Stone]{cui2025towards}
Jiaxun Cui, Chen Tang, Jarrett Holtz, Janice Nguyen, Alessandro~G Allievi, Hang Qiu, and Peter Stone.
\newblock Towards natural language communication for cooperative autonomous driving via self-play.
\newblock \emph{arXiv preprint arXiv:2505.18334}, 2025{\natexlab{a}}.

\bibitem[Cui et~al.(2025{\natexlab{b}})Cui, Fang, Hang, and Sun]{cui2025vehicle}
Yiming Cui, Shiyu Fang, Peng Hang, and Jian Sun.
\newblock A vehicle-infrastructure multi-layer cooperative decision-making framework.
\newblock \emph{arXiv preprint arXiv:2503.16552}, 2025{\natexlab{b}}.

\bibitem[Dauner et~al.(2024)Dauner, Hallgarten, Li, Weng, Huang, Yang, Li, Gilitschenski, Ivanovic, Pavone, et~al.]{dauner2024navsim}
Daniel Dauner, Marcel Hallgarten, Tianyu Li, Xinshuo Weng, Zhiyu Huang, Zetong Yang, Hongyang Li, Igor Gilitschenski, Boris Ivanovic, Marco Pavone, et~al.
\newblock Navsim: Data-driven non-reactive autonomous vehicle simulation and benchmarking.
\newblock \emph{Advances in Neural Information Processing Systems}, 37:\penalty0 28706--28719, 2024.

\bibitem[Dosovitskiy et~al.(2017)Dosovitskiy, Ros, Codevilla, Lopez, and Koltun]{dosovitskiy2017carla}
Alexey Dosovitskiy, German Ros, Felipe Codevilla, Antonio Lopez, and Vladlen Koltun.
\newblock Carla: An open urban driving simulator.
\newblock In \emph{Conference on robot learning}, pages 1--16. PMLR, 2017.

\bibitem[Fan et~al.(2024)Fan, Yu, Yang, Yuan, and Nie]{fan2024quest}
Siqi Fan, Haibao Yu, Wenxian Yang, Jirui Yuan, and Zaiqing Nie.
\newblock Quest: Query stream for practical cooperative perception.
\newblock In \emph{2024 IEEE International Conference on Robotics and Automation (ICRA)}, pages 18436--18442. IEEE, 2024.

\bibitem[Feng et~al.(2024)Feng, Wang, Han, Zhang, and Zhu]{feng2024u2udata}
Tongtong Feng, Xin Wang, Feilin Han, Leping Zhang, and Wenwu Zhu.
\newblock U2udata: A large-scale cooperative perception dataset for swarm uavs autonomous flight.
\newblock In \emph{Proceedings of the 32nd ACM International Conference on Multimedia}, pages 7600--7608, 2024.

\bibitem[Fu et~al.(2025)Fu, Yang, Ma, and Zhang]{fu2025self}
Fuji Fu, Jinfu Yang, Jiaqi Ma, and Jiahui Zhang.
\newblock Self-supervised visual odometry based on scene appearance-structure incremental fusion.
\newblock \emph{IEEE Transactions on Intelligent Transportation Systems}, 2025.

\bibitem[Gao et~al.(2025{\natexlab{a}})Gao, Wu, Zhang, Tian, Zhou, and Tu]{gao2025automated}
Xiangbo Gao, Keshu Wu, Hao Zhang, Kexin Tian, Yang Zhou, and Zhengzhong Tu.
\newblock Automated vehicles should be connected with natural language.
\newblock \emph{arXiv preprint arXiv:2507.01059}, 2025{\natexlab{a}}.

\bibitem[Gao et~al.(2025{\natexlab{b}})Gao, Wu, Wang, Liu, Zhou, and Tu]{gao2025langcoop}
Xiangbo Gao, Yuheng Wu, Rujia Wang, Chenxi Liu, Yang Zhou, and Zhengzhong Tu.
\newblock Langcoop: Collaborative driving with language.
\newblock In \emph{Proceedings of the Computer Vision and Pattern Recognition Conference}, pages 4226--4237, 2025{\natexlab{b}}.

\bibitem[Gao et~al.(2025{\natexlab{c}})Gao, Wu, Yang, Luo, Wu, Chen, Wang, Liu, Zhou, and Tu]{gao2025airv2x}
Xiangbo Gao, Yuheng Wu, Fengze Yang, Xuewen Luo, Keshu Wu, Xinghao Chen, Yuping Wang, Chenxi Liu, Yang Zhou, and Zhengzhong Tu.
\newblock Airv2x: Unified air-ground vehicle-to-everything collaboration.
\newblock \emph{arXiv preprint arXiv:2506.19283}, 2025{\natexlab{c}}.

\bibitem[Gao et~al.(2025{\natexlab{d}})Gao, Xu, Li, Wang, Fan, and Tu]{gao2025stamp}
Xiangbo Gao, Runsheng Xu, Jiachen Li, Ziran Wang, Zhiwen Fan, and Zhengzhong Tu.
\newblock Stamp: Scalable task and model-agnostic collaborative perception.
\newblock \emph{arXiv preprint arXiv:2501.18616}, 2025{\natexlab{d}}.

\bibitem[Gularte et~al.(2024)Gularte, Da~Costa, Vargas, Da~Silva, Santos, Wang, M{\"u}ller, Lipps, de~Sousa~J{\'u}nior, de~Britto Vidal~Filho, et~al.]{gularte2024integrating}
Kevin Herman~Muraro Gularte, Jo{\~a}o Paulo~Javidi Da~Costa, Jos{\'e} Alfredo~Ruiz Vargas, Antonio~Santos Da~Silva, Giovanni~Almeida Santos, Yuming Wang, Christian~Alfons M{\"u}ller, Christoph Lipps, Rafael~Tim{\'o}teo de Sousa~J{\'u}nior, Walter de Britto Vidal~Filho, et~al.
\newblock Integrating cybersecurity in v2x: A review of simulation environments.
\newblock \emph{IEEE Access}, 2024.

\bibitem[Hao et~al.(2024)Hao, Fan, Dai, Zhang, Li, Wang, Yu, Yang, Yuan, and Nie]{hao2024rcooper}
Ruiyang Hao, Siqi Fan, Yingru Dai, Zhenlin Zhang, Chenxi Li, Yuntian Wang, Haibao Yu, Wenxian Yang, Jirui Yuan, and Zaiqing Nie.
\newblock Rcooper: A real-world large-scale dataset for roadside cooperative perception.
\newblock In \emph{Proceedings of the IEEE/CVF conference on computer vision and pattern recognition}, pages 22347--22357, 2024.

\bibitem[Hao et~al.(2025)Hao, Jing, Yu, and Nie]{hao2025styledrive}
Ruiyang Hao, Bowen Jing, Haibao Yu, and Zaiqing Nie.
\newblock Styledrive: Towards driving-style aware benchmarking of end-to-end autonomous driving.
\newblock \emph{arXiv preprint arXiv:2506.23982}, 2025.

\bibitem[Hou et~al.(2025)Hou, Zou, Zhang, Chen, Yang, Zhang, Zhuo, Chen, Chen, and Ma]{hou2025agc}
Yunhao Hou, Bochao Zou, Min Zhang, Ran Chen, Shangdong Yang, Yanmei Zhang, Junbao Zhuo, Siheng Chen, Jiansheng Chen, and Huimin Ma.
\newblock Agc-drive: A large-scale dataset for real-world aerial-ground collaboration in driving scenarios.
\newblock \emph{arXiv preprint arXiv:2506.16371}, 2025.

\bibitem[Hu et~al.(2023)Hu, Yang, Chen, Li, Sima, Zhu, Chai, Du, Lin, Wang, et~al.]{hu2023planning}
Yihan Hu, Jiazhi Yang, Li Chen, Keyu Li, Chonghao Sima, Xizhou Zhu, Siqi Chai, Senyao Du, Tianwei Lin, Wenhai Wang, et~al.
\newblock Planning-oriented autonomous driving.
\newblock In \emph{Proceedings of the IEEE/CVF conference on computer vision and pattern recognition}, pages 17853--17862, 2023.

\bibitem[Huang et~al.(2023{\natexlab{a}})Huang, Liu, and Lv]{huang2023gameformer}
Zhiyu Huang, Haochen Liu, and Chen Lv.
\newblock Gameformer: Game-theoretic modeling and learning of transformer-based interactive prediction and planning for autonomous driving.
\newblock In \emph{Proceedings of the IEEE/CVF International Conference on Computer Vision}, pages 3903--3913, 2023{\natexlab{a}}.

\bibitem[Huang et~al.(2023{\natexlab{b}})Huang, Liu, Wu, and Lv]{huang2023differentiable}
Zhiyu Huang, Haochen Liu, Jingda Wu, and Chen Lv.
\newblock Differentiable integrated motion prediction and planning with learnable cost function for autonomous driving.
\newblock \emph{IEEE transactions on neural networks and learning systems}, 35\penalty0 (11):\penalty0 15222--15236, 2023{\natexlab{b}}.

\bibitem[Hui et~al.(2023)Hui, Hu, Cheng, Zhao, Chen, Luan, and Aldubaikhy]{hui2023rcfl}
Yilong Hui, Jie Hu, Nan Cheng, Gaosheng Zhao, Rui Chen, Tom~H Luan, and Khalid Aldubaikhy.
\newblock Rcfl: Redundancy-aware collaborative federated learning in vehicular networks.
\newblock \emph{IEEE Transactions on Intelligent Transportation Systems}, 25\penalty0 (6):\penalty0 5539--5553, 2023.

\bibitem[Jia et~al.(2024)Jia, Yang, Li, Zhang, and Yan]{jia2024bench2drive}
Xiaosong Jia, Zhenjie Yang, Qifeng Li, Zhiyuan Zhang, and Junchi Yan.
\newblock Bench2drive: Towards multi-ability benchmarking of closed-loop end-to-end autonomous driving.
\newblock \emph{Advances in Neural Information Processing Systems}, 37:\penalty0 819--844, 2024.

\bibitem[Jiang et~al.(2023)Jiang, Chen, Xu, Liao, Chen, Zhou, Zhang, Liu, Huang, and Wang]{jiang2023vad}
Bo Jiang, Shaoyu Chen, Qing Xu, Bencheng Liao, Jiajie Chen, Helong Zhou, Qian Zhang, Wenyu Liu, Chang Huang, and Xinggang Wang.
\newblock Vad: Vectorized scene representation for efficient autonomous driving.
\newblock In \emph{Proceedings of the IEEE/CVF International Conference on Computer Vision}, pages 8340--8350, 2023.

\bibitem[Jiang et~al.(2025)Jiang, Huang, Qian, Luo, Zhu, Zhong, Tang, Kong, Wang, Jiao, et~al.]{jiang2025survey}
Sicong Jiang, Zilin Huang, Kangan Qian, Ziang Luo, Tianze Zhu, Yang Zhong, Yihong Tang, Menglin Kong, Yunlong Wang, Siwen Jiao, et~al.
\newblock A survey on vision-language-action models for autonomous driving.
\newblock \emph{arXiv preprint arXiv:2506.24044}, 2025.

\bibitem[Kong et~al.(2023)Kong, Jiang, Jia, Shi, Xu, and Liu]{kong2023dusa}
Xianghao Kong, Wentao Jiang, Jinrang Jia, Yifeng Shi, Runsheng Xu, and Si Liu.
\newblock Dusa: Decoupled unsupervised sim2real adaptation for vehicle-to-everything collaborative perception.
\newblock In \emph{Proceedings of the 31st ACM International Conference on Multimedia}, pages 1943--1954, 2023.

\bibitem[Lei et~al.(2025)Lei, Zhou, Li, Ma, and Hu]{lei2025risk}
Mingyue Lei, Zewei Zhou, Hongchen Li, Jiaqi Ma, and Jia Hu.
\newblock Risk map as middleware: Towards interpretable cooperative end-to-end autonomous driving for risk-aware planning.
\newblock \emph{arXiv preprint arXiv:2508.07686}, 2025.

\bibitem[Li et~al.(2024{\natexlab{a}})Li, Li, Liu, Xu, Tu, Guo, Li, and Yu]{li2024v2x}
Baolu Li, Jinlong Li, Xinyu Liu, Runsheng Xu, Zhengzhong Tu, Jiacheng Guo, Xiaopeng Li, and Hongkai Yu.
\newblock V2x-dgw: Domain generalization for multi-agent perception under adverse weather conditions.
\newblock \emph{arXiv preprint arXiv:2403.11371}, 2024{\natexlab{a}}.

\bibitem[Li et~al.(2024{\natexlab{b}})Li, Zhang, Zhou, Cai, and Yu]{li2024coarse}
Hanlei Li, Guangyi Zhang, Kequan Zhou, Yunlong Cai, and Guanding Yu.
\newblock Coarse-to-fine: A dual-phase channel-adaptive method for wireless image transmission.
\newblock \emph{arXiv preprint arXiv:2412.08211}, 2024{\natexlab{b}}.

\bibitem[Li et~al.(2025)Li, Ma, Chang, He, and Huang]{li2025efficient}
Wei Li, Lin Ma, Haoze Chang, Xiangyun He, and Longteng Huang.
\newblock Efficient collaborative perception with integrated uncertainty estimation via evidence regression.
\newblock \emph{IEEE Transactions on Intelligent Transportation Systems}, 2025.

\bibitem[Li et~al.(2024{\natexlab{c}})Li, Yin, Li, Xu, Yang, and Shen]{li2024di}
Xiang Li, Junbo Yin, Wei Li, Chengzhong Xu, Ruigang Yang, and Jianbing Shen.
\newblock Di-v2x: Learning domain-invariant representation for vehicle-infrastructure collaborative 3d object detection.
\newblock In \emph{Proceedings of the AAAI Conference on Artificial Intelligence}, pages 3208--3215, 2024{\natexlab{c}}.

\bibitem[Li et~al.(2022)Li, Ma, An, Wang, Zhong, Chen, and Feng]{li2022v2x}
Yiming Li, Dekun Ma, Ziyan An, Zixun Wang, Yiqi Zhong, Siheng Chen, and Chen Feng.
\newblock V2x-sim: Multi-agent collaborative perception dataset and benchmark for autonomous driving.
\newblock \emph{IEEE Robotics and Automation Letters}, 7\penalty0 (4):\penalty0 10914--10921, 2022.

\bibitem[Luo et~al.(2024)Luo, Wang, Zhang, and Wu]{luo2024full}
Xiaoqing Luo, Juan Wang, Zhancheng Zhang, and Xiao-jun Wu.
\newblock A full-scale hierarchical encoder-decoder network with cascading edge-prior for infrared and visible image fusion.
\newblock \emph{Pattern Recognition}, 148:\penalty0 110192, 2024.

\bibitem[Luo et~al.(2025)Luo, Yang, Ding, Gao, Xing, Zhou, Tu, and Liu]{luo2025v2x}
Xuewen Luo, Fengze Yang, Fan Ding, Xiangbo Gao, Shuo Xing, Yang Zhou, Zhengzhong Tu, and Chenxi Liu.
\newblock V2x-unipool: Unifying multimodal perception and knowledge reasoning for autonomous driving.
\newblock \emph{arXiv preprint arXiv:2506.02580}, 2025.

\bibitem[Ma et~al.(2024)Ma, Xu, Liu, and Zhang]{ma2024class}
Xi-Ao Ma, Hao Xu, Yi Liu, and Justin~Zuopeng Zhang.
\newblock Class-specific feature selection using fuzzy information-theoretic metrics.
\newblock \emph{Engineering Applications of Artificial Intelligence}, 136:\penalty0 109035, 2024.

\bibitem[Mouawad and Mannoni(2021)]{mouawad2021collective}
Nadia Mouawad and Val{\'e}rian Mannoni.
\newblock Collective perception messages: New low complexity fusion and v2x connectivity analysis.
\newblock In \emph{2021 IEEE 94th Vehicular Technology Conference (VTC2021-Fall)}, pages 1--5. IEEE, 2021.

\bibitem[Pan et~al.(2024)Pan, Yaman, Nesti, Mallik, Allievi, Velipasalar, and Ren]{pan2024vlp}
Chenbin Pan, Burhaneddin Yaman, Tommaso Nesti, Abhirup Mallik, Alessandro~G Allievi, Senem Velipasalar, and Liu Ren.
\newblock Vlp: Vision language planning for autonomous driving.
\newblock In \emph{Proceedings of the IEEE/CVF Conference on Computer Vision and Pattern Recognition}, pages 14760--14769, 2024.

\bibitem[Peng et~al.(2023)Peng, Genova, Jiang, Tagliasacchi, Pollefeys, Funkhouser, et~al.]{peng2023openscene}
Songyou Peng, Kyle Genova, Chiyu Jiang, Andrea Tagliasacchi, Marc Pollefeys, Thomas Funkhouser, et~al.
\newblock Openscene: 3d scene understanding with open vocabularies.
\newblock In \emph{Proceedings of the IEEE/CVF conference on computer vision and pattern recognition}, pages 815--824, 2023.

\bibitem[Ren et~al.(2024)Ren, Lei, Wang, Dianati, Wang, Chen, and Zhang]{ren2024interruption}
Shunli Ren, Zixing Lei, Zi Wang, Mehrdad Dianati, Yafei Wang, Siheng Chen, and Wenjun Zhang.
\newblock Interruption-aware cooperative perception for v2x communication-aided autonomous driving.
\newblock \emph{IEEE Transactions on Intelligent Vehicles}, 9\penalty0 (4):\penalty0 4698--4714, 2024.

\bibitem[Song et~al.(2024{\natexlab{a}})Song, Festag, Jagtap, Bialdyga, Yan, Otte, Sadashivaiah, and Knoll]{10588500}
Rui Song, Andreas Festag, Abhishek~Dinkar Jagtap, Maximilian Bialdyga, Zhiran Yan, Maximilian Otte, Sanath~Tiptur Sadashivaiah, and Alois Knoll.
\newblock First mile: An open innovation lab for infrastructure-assisted cooperative intelligent transportation systems.
\newblock In \emph{2024 IEEE Intelligent Vehicles Symposium (IV)}, pages 1635--1642, 2024{\natexlab{a}}.

\bibitem[Song et~al.(2024{\natexlab{b}})Song, Liang, Cao, Yan, Zimmer, Gross, Festag, and Knoll]{Song_2024_CVPR}
Rui Song, Chenwei Liang, Hu Cao, Zhiran Yan, Walter Zimmer, Markus Gross, Andreas Festag, and Alois Knoll.
\newblock Collaborative semantic occupancy prediction with hybrid feature fusion in connected automated vehicles.
\newblock In \emph{Proceedings of the IEEE/CVF Conference on Computer Vision and Pattern Recognition (CVPR)}, pages 17996--18006, 2024{\natexlab{b}}.

\bibitem[Souri et~al.(2024)Souri, Zarei, Hemmati, and Gao]{souri2024systematic}
Alireza Souri, Mani Zarei, Atefeh Hemmati, and Mingliang Gao.
\newblock A systematic literature review of vehicular connectivity and v2x communications: Technical aspects and new challenges.
\newblock \emph{International Journal of Communication Systems}, 37\penalty0 (10):\penalty0 e5780, 2024.

\bibitem[Sumner et~al.(2013)Sumner, Eisenhart, Baker, et~al.]{sumner2013sae}
Roy Sumner, Bruce Eisenhart, John Baker, et~al.
\newblock Sae j2735 standard: applying the systems engineering process.
\newblock Technical report, United States. Department of Transportation. Intelligent Transportation~…, 2013.

\bibitem[Sun et~al.(2020)Sun, Kretzschmar, Dotiwalla, Chouard, Patnaik, Tsui, Guo, Zhou, Chai, Caine, et~al.]{sun2020scalability}
Pei Sun, Henrik Kretzschmar, Xerxes Dotiwalla, Aurelien Chouard, Vijaysai Patnaik, Paul Tsui, James Guo, Yin Zhou, Yuning Chai, Benjamin Caine, et~al.
\newblock Scalability in perception for autonomous driving: Waymo open dataset.
\newblock In \emph{Proceedings of the IEEE/CVF conference on computer vision and pattern recognition}, pages 2446--2454, 2020.

\bibitem[Tan et~al.(2024)Tan, Lyu, Li, Hu, Feng, Xu, Zhang, Yao, and Wang]{tan2024dynamic}
Jiayao Tan, Fan Lyu, Linyan Li, Fuyuan Hu, Tingliang Feng, Fenglei Xu, Zhang Zhang, Rui Yao, and Liang Wang.
\newblock Dynamic v2x perception from road-to-vehicle vision.
\newblock \emph{IEEE Transactions on Intelligent Vehicles}, 2024.

\bibitem[Vogt(2024)]{vogt2024comprehensive}
Jonas Vogt.
\newblock A comprehensive overview of the protocols associated with intelligent transportation systems.
\newblock \emph{arXiv preprint arXiv:2407.12799}, 2024.

\bibitem[Wang et~al.(2025{\natexlab{a}})Wang, Cao, Zhong, Zhang, Yu, He, and Xu]{wang2025griffin}
Jiahao Wang, Xiangyu Cao, Jiaru Zhong, Yuner Zhang, Haibao Yu, Lei He, and Shaobing Xu.
\newblock Griffin: Aerial-ground cooperative detection and tracking dataset and benchmark.
\newblock \emph{arXiv preprint arXiv:2503.06983}, 2025{\natexlab{a}}.

\bibitem[Wang et~al.(2025{\natexlab{b}})Wang, Yuan, Zhang, He, Xu, and Wang]{wang2025v2x}
Sichao Wang, Ming Yuan, Chuang Zhang, Lei He, Qing Xu, and Jianqiang Wang.
\newblock V2x-dgpe: Addressing domain gaps and pose errors for robust collaborative 3d object detection.
\newblock In \emph{2025 IEEE Intelligent Vehicles Symposium (IV)}, pages 2074--2080. IEEE, 2025{\natexlab{b}}.

\bibitem[Wang et~al.(2024)Wang, He, Fan, Li, Chen, and Zhang]{wang2024driving}
Yuqi Wang, Jiawei He, Lue Fan, Hongxin Li, Yuntao Chen, and Zhaoxiang Zhang.
\newblock Driving into the future: Multiview visual forecasting and planning with world model for autonomous driving.
\newblock In \emph{Proceedings of the IEEE/CVF Conference on Computer Vision and Pattern Recognition}, pages 14749--14759, 2024.

\bibitem[Wang et~al.(2025{\natexlab{c}})Wang, Sun, Sun, Wang, Li, Zhao, Wu, Liang, Yin, Wang, et~al.]{wang2025toward}
Yixian Wang, Geng Sun, Zemin Sun, Jiacheng Wang, Jiahui Li, Changyuan Zhao, Jing Wu, Shuang Liang, Minghao Yin, Pengfei Wang, et~al.
\newblock Toward realization of low-altitude economy networks: Core architecture, integrated technologies, and future directions.
\newblock \emph{arXiv preprint arXiv:2504.21583}, 2025{\natexlab{c}}.

\bibitem[Wang et~al.(2025{\natexlab{d}})Wang, Xing, Can, Li, Hua, Tian, Mo, Gao, Wu, Zhou, et~al.]{wang2025generative}
Yuping Wang, Shuo Xing, Cui Can, Renjie Li, Hongyuan Hua, Kexin Tian, Zhaobin Mo, Xiangbo Gao, Keshu Wu, Sulong Zhou, et~al.
\newblock Generative ai for autonomous driving: Frontiers and opportunities.
\newblock \emph{arXiv preprint arXiv:2505.08854}, 2025{\natexlab{d}}.

\bibitem[Wang et~al.(2025{\natexlab{e}})Wang, Xu, Zhuang, Xu, Wang, Liu, Chen, and Zhang]{wang2025coopdetr}
Zhe Wang, Shaocong Xu, Xucai Zhuang, Tongda Xu, Yan Wang, Jingjing Liu, Yilun Chen, and Ya-Qin Zhang.
\newblock Coopdetr: A unified cooperative perception framework for 3d detection via object query.
\newblock \emph{arXiv preprint arXiv:2502.19313}, 2025{\natexlab{e}}.

\bibitem[Wei et~al.(2025)Wei, Qin, Zimmer, Wu, and Barth]{wei2025hecofuse}
Chuheng Wei, Ziye Qin, Walter Zimmer, Guoyuan Wu, and Matthew~J Barth.
\newblock Hecofuse: Cross-modal complementary v2x cooperative perception with heterogeneous sensors.
\newblock \emph{arXiv preprint arXiv:2507.13677}, 2025.

\bibitem[Wu et~al.(2025)Wu, Li, Zhou, Gan, You, Cheng, Zhu, Parker, Ran, Noyce, et~al.]{wu2025v2x}
Keshu Wu, Pei Li, Yang Zhou, Rui Gan, Junwei You, Yang Cheng, Jingwen Zhu, Steven~T Parker, Bin Ran, David~A Noyce, et~al.
\newblock V2x-llm: Enhancing v2x integration and understanding in connected vehicle corridors.
\newblock \emph{arXiv preprint arXiv:2503.02239}, 2025.

\bibitem[Xia et~al.(2025)Xia, Yuan, Luo, Fu, Li, Zhu, Luo, Chen, and Li]{xia2025one}
Yuchen Xia, Quan Yuan, Guiyang Luo, Xiaoyuan Fu, Yang Li, Xuanhan Zhu, Tianyou Luo, Siheng Chen, and Jinglin Li.
\newblock One is plenty: A polymorphic feature interpreter for immutable heterogeneous collaborative perception.
\newblock In \emph{Proceedings of the Computer Vision and Pattern Recognition Conference}, pages 1592--1601, 2025.

\bibitem[Xiang et~al.(2024)Xiang, Zheng, Xia, Xu, Gao, Zhou, Han, Ji, Li, Meng, et~al.]{xiang2024v2x}
Hao Xiang, Zhaoliang Zheng, Xin Xia, Runsheng Xu, Letian Gao, Zewei Zhou, Xu Han, Xinkai Ji, Mingxi Li, Zonglin Meng, et~al.
\newblock V2x-real: a largs-scale dataset for vehicle-to-everything cooperative perception.
\newblock In \emph{European Conference on Computer Vision}, pages 455--470. Springer, 2024.

\bibitem[Xiang et~al.(2025)Xiang, Zheng, Xia, Zhao, Gao, Zhou, Cai, Zhang, and Ma]{xiang2025v2x}
Hao Xiang, Zhaoliang Zheng, Xin Xia, Seth~Z Zhao, Letian Gao, Zewei Zhou, Tianhui Cai, Yun Zhang, and Jiaqi Ma.
\newblock V2x-realo: An open online framework and dataset for cooperative perception in reality.
\newblock \emph{arXiv preprint arXiv:2503.10034}, 2025.

\bibitem[Xing et~al.(2024)Xing, Hua, Gao, Zhu, Li, Tian, Li, Huang, Yang, Wang, et~al.]{xing2024autotrust}
Shuo Xing, Hongyuan Hua, Xiangbo Gao, Shenzhe Zhu, Renjie Li, Kexin Tian, Xiaopeng Li, Heng Huang, Tianbao Yang, Zhangyang Wang, et~al.
\newblock Autotrust: Benchmarking trustworthiness in large vision language models for autonomous driving.
\newblock \emph{arXiv preprint arXiv:2412.15206}, 2024.

\bibitem[Xu et~al.(2024)Xu, Chen, Zheng, and Feng]{xu2024delay}
Fan Xu, Chen Chen, Haifeng Zheng, and Xinxin Feng.
\newblock Delay-aware cooperative perception with deep reinforcement learning in vehicular networks.
\newblock In \emph{2024 9th International Conference on Computer and Communication Systems (ICCCS)}, pages 980--985. IEEE, 2024.

\bibitem[Xu et~al.(2022)Xu, Xiang, Tu, Xia, Yang, and Ma]{xu2022v2x}
Runsheng Xu, Hao Xiang, Zhengzhong Tu, Xin Xia, Ming-Hsuan Yang, and Jiaqi Ma.
\newblock V2x-vit: Vehicle-to-everything cooperative perception with vision transformer.
\newblock In \emph{European conference on computer vision}, pages 107--124. Springer, 2022.

\bibitem[Xu et~al.(2023{\natexlab{a}})Xu, Xia, Li, Li, Zhang, Tu, Meng, Xiang, Dong, Song, et~al.]{xu2023v2v4real}
Runsheng Xu, Xin Xia, Jinlong Li, Hanzhao Li, Shuo Zhang, Zhengzhong Tu, Zonglin Meng, Hao Xiang, Xiaoyu Dong, Rui Song, et~al.
\newblock V2v4real: A real-world large-scale dataset for vehicle-to-vehicle cooperative perception.
\newblock In \emph{Proceedings of the IEEE/CVF conference on computer vision and pattern recognition}, pages 13712--13722, 2023{\natexlab{a}}.

\bibitem[Xu et~al.(2025)Xu, Chen, Tu, and Yang]{10715696}
Runsheng Xu, Chia-Ju Chen, Zhengzhong Tu, and Ming-Hsuan Yang.
\newblock V2x-vitv2: Improved vision transformers for vehicle-to-everything cooperative perception.
\newblock \emph{IEEE Transactions on Pattern Analysis and Machine Intelligence}, 47\penalty0 (1):\penalty0 650--662, 2025.

\bibitem[Xu et~al.(2023{\natexlab{b}})Xu, Liu, Dai, Xie, Cao, and Luo]{xu2023cooperative}
Xincao Xu, Kai Liu, Penglin Dai, Ruitao Xie, Jingjing Cao, and Jiangtao Luo.
\newblock Cooperative sensing and heterogeneous information fusion in vcps: A multi-agent deep reinforcement learning approach.
\newblock \emph{IEEE Transactions on Intelligent Transportation Systems}, 25\penalty0 (6):\penalty0 4876--4891, 2023{\natexlab{b}}.

\bibitem[Yang et~al.(2022)Yang, Shi, Xing, and Liu]{yang2022autonomous}
Xun Yang, Yunyang Shi, Jiping Xing, and Zhiyuan Liu.
\newblock Autonomous driving under v2x environment: state-of-the-art survey and challenges.
\newblock \emph{Intelligent Transportation Infrastructure}, 1:\penalty0 liac020, 2022.

\bibitem[Yi et~al.(2024)Yi, Zhang, and Liu]{10720085}
Sheng Yi, Hao Zhang, and Kai Liu.
\newblock V2iviewer: Towards efficient collaborative perception via point cloud data fusion and vehicle-to-infrastructure communications.
\newblock \emph{IEEE Transactions on Network Science and Engineering}, 11\penalty0 (6):\penalty0 6219--6230, 2024.

\bibitem[You et~al.(2024)You, Shi, Jiang, Huang, Gan, Wu, Cheng, Li, and Ran]{you2024v2x}
Junwei You, Haotian Shi, Zhuoyu Jiang, Zilin Huang, Rui Gan, Keshu Wu, Xi Cheng, Xiaopeng Li, and Bin Ran.
\newblock V2x-vlm: End-to-end v2x cooperative autonomous driving through large vision-language models.
\newblock \emph{arXiv preprint arXiv:2408.09251}, 2024.

\bibitem[Yu et~al.(2022)Yu, Luo, Shu, Huo, Yang, Shi, Guo, Li, Hu, Yuan, et~al.]{yu2022dair}
Haibao Yu, Yizhen Luo, Mao Shu, Yiyi Huo, Zebang Yang, Yifeng Shi, Zhenglong Guo, Hanyu Li, Xing Hu, Jirui Yuan, et~al.
\newblock Dair-v2x: A large-scale dataset for vehicle-infrastructure cooperative 3d object detection.
\newblock In \emph{Proceedings of the IEEE/CVF conference on computer vision and pattern recognition}, pages 21361--21370, 2022.

\bibitem[Yu et~al.(2023{\natexlab{a}})Yu, Tang, Xie, Mao, Luo, and Nie]{yu2023flow}
Haibao Yu, Yingjuan Tang, Enze Xie, Jilei Mao, Ping Luo, and Zaiqing Nie.
\newblock Flow-based feature fusion for vehicle-infrastructure cooperative 3d object detection.
\newblock \emph{Advances in Neural Information Processing Systems}, 36:\penalty0 34493--34503, 2023{\natexlab{a}}.

\bibitem[Yu et~al.(2023{\natexlab{b}})Yu, Yang, Ruan, Yang, Tang, Gao, Hao, Shi, Pan, Sun, et~al.]{yu2023v2x}
Haibao Yu, Wenxian Yang, Hongzhi Ruan, Zhenwei Yang, Yingjuan Tang, Xu Gao, Xin Hao, Yifeng Shi, Yifeng Pan, Ning Sun, et~al.
\newblock V2x-seq: A large-scale sequential dataset for vehicle-infrastructure cooperative perception and forecasting.
\newblock In \emph{Proceedings of the IEEE/CVF Conference on Computer Vision and Pattern Recognition}, pages 5486--5495, 2023{\natexlab{b}}.

\bibitem[Yu et~al.(2025)Yu, Yang, Zhong, Yang, Fan, Luo, and Nie]{yu2025end}
Haibao Yu, Wenxian Yang, Jiaru Zhong, Zhenwei Yang, Siqi Fan, Ping Luo, and Zaiqing Nie.
\newblock End-to-end autonomous driving through v2x cooperation.
\newblock In \emph{Proceedings of the AAAI Conference on Artificial Intelligence}, pages 9598--9606, 2025.

\bibitem[Yusuf et~al.(2024)Yusuf, Khan, and Souissi]{yusuf2024vehicle}
Syed~Adnan Yusuf, Arshad Khan, and Riad Souissi.
\newblock Vehicle-to-everything (v2x) in the autonomous vehicles domain--a technical review of communication, sensor, and ai technologies for road user safety.
\newblock \emph{Transportation Research Interdisciplinary Perspectives}, 23:\penalty0 100980, 2024.

\bibitem[Zha et~al.(2025)Zha, Shangguan, Chen, Chai, Qiu, and L{\'o}pez]{zha2025heterogeneous}
Yuanyuan Zha, Wei Shangguan, Junjie Chen, Linguo Chai, Weizhi Qiu, and Antonio~M L{\'o}pez.
\newblock Heterogeneous multiscale cooperative perception for connected autonomous vehicles via v2x interaction.
\newblock \emph{IEEE Internet of Things Journal}, 2025.

\bibitem[Zhao et~al.(2024{\natexlab{a}})Zhao, Nybacka, Aramrattana, Rothh{\"a}mel, Habibovic, Drugge, and Jiang]{zhao2024remote}
Lin Zhao, Mikael Nybacka, Maytheewat Aramrattana, Malte Rothh{\"a}mel, Azra Habibovic, Lars Drugge, and Frank Jiang.
\newblock Remote driving of road vehicles: A survey of driving feedback, latency, support control, and real applications.
\newblock \emph{IEEE Transactions on Intelligent Vehicles}, 2024{\natexlab{a}}.

\bibitem[Zhao et~al.(2024{\natexlab{b}})Zhao, Xiang, Xu, Xia, Zhou, and Ma]{zhao2024coopre}
Seth~Z Zhao, Hao Xiang, Chenfeng Xu, Xin Xia, Bolei Zhou, and Jiaqi Ma.
\newblock Coopre: Cooperative pretraining for v2x cooperative perception.
\newblock \emph{arXiv preprint arXiv:2408.11241}, 2024{\natexlab{b}}.

\bibitem[Zhong et~al.(2024)Zhong, Yu, Zhu, Xu, Yang, Nie, and Sun]{zhong2024leveraging}
Jiaru Zhong, Haibao Yu, Tianyi Zhu, Jiahui Xu, Wenxian Yang, Zaiqing Nie, and Chao Sun.
\newblock Leveraging temporal contexts to enhance vehicle-infrastructure cooperative perception.
\newblock In \emph{2024 IEEE 27th International Conference on Intelligent Transportation Systems (ITSC)}, pages 915--922. IEEE, 2024.

\bibitem[Zhong et~al.(2025)Zhong, Wang, Xu, Li, Nie, and Yu]{zhong2025cooptrack}
Jiaru Zhong, Jiahao Wang, Jiahui Xu, Xiaofan Li, Zaiqing Nie, and Haibao Yu.
\newblock Cooptrack: Exploring end-to-end learning for efficient cooperative sequential perception.
\newblock \emph{arXiv preprint arXiv:2507.19239}, 2025.

\bibitem[Zimmer et~al.(2024)Zimmer, Wardana, Sritharan, Zhou, Song, and Knoll]{zimmer2024tumtraf}
Walter Zimmer, Gerhard~Arya Wardana, Suren Sritharan, Xingcheng Zhou, Rui Song, and Alois~C Knoll.
\newblock Tumtraf v2x cooperative perception dataset.
\newblock In \emph{Proceedings of the IEEE/CVF conference on computer vision and pattern recognition}, pages 22668--22677, 2024.

\end{thebibliography}
}

\end{document}